# Code Generation for High-Level Synthesis of Multiresolution Applications on FPGAs

Moritz Schmid, Oliver Reiche, Christian Schmitt, Frank Hannig, and Jürgen Teich
Hardware/Software Co-Design, Department of Computer Science
Friedrich-Alexander-Universität Erlangen-Nürnberg (FAU), Germany.

*Abstract*— Multiresolution Analysis (MRA) is a mathematical method that is based on working on a problem at different scales. One of its applications is medical imaging where processing at multiple scales—based on the concept of Gaussian and Laplacian image pyramids—is a well-known technique. It is often applied to reduce noise while preserving image detail on different levels of granularity without modifying the filter kernel. In scientific computing, multigrid methods are a popular choice, as they are asymptotically optimal solvers for elliptic Partial Differential Equations (PDEs). As such algorithms have a very high computational complexity that would overwhelm CPUs in the presence of real-time constraints, application-specific processors come into consideration for implementation. Despite of huge advancements in leveraging productivity in the respective fields, designers are still required to have detailed knowledge about coding techniques and the targeted architecture to achieve efficient solutions. Recently, the HIPA$^{cc}$ framework was proposed as a means for automatic code generation of image processing algorithms, based on a Domain-Specific Language (DSL). From the same code base, it is possible to generate code for efficient implementations on several accelerator technologies including different types of Graphics Processing Units (GPUs) as well as reconfigurable logic (FPGAs). In this work, we demonstrate the ability of HIPA$^{cc}$ to generate code for the implementation of multiresolution applications on FPGAs and embedded GPUs.

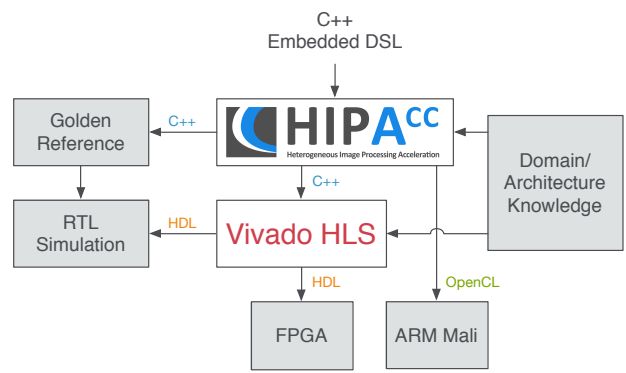

Figure 1. Design flow of the combination of HIPA$^{cc}$ and Vivado HLS.

## I. Introduction

A few among numerous applications of MRA are signal detection, differential equation solving, information retrieval, computer vision, as well as signal and image processing. The algorithms used to solve problems in industry and scientific computing are becoming more and more complex and must deliver enough performance to process vast amounts of data often under rigid resource and energy constraints. Due to these requirements, hardware accelerators, such as embedded GPUs and Field Programmable Gate Arrays (FPGAs) are ideal targets for the implementation. Although there has been tremendous progress in making the respective programming models more approachable, a deep understanding of the algorithmic details and the hardware architecture are necessary to achieve good results. To ease the burden on developers, DSLs aim at combining architecture- and domain-specific knowledge, thereby delivering performance, productivity, and portability. So far, DSLs have been researched for a long time for Central Processing Units (CPUs) as well as GPUs, and recently have also targeted hardware design [15, 7], which has mostly been the prime domain for High-Level Synthesis (HLS). Over the past decades, C-based HLS focusing on FPGAs has become very sophisticated, producing designs that can rival hand-coded Register-Transfer Level (RTL). A drawback is that these frameworks must be very flexible and although being able to create an efficient hardware design from a C-based language can significantly shorten the development time, architectural knowledge and specific coding techniques are still a must. A remedy to this situation is to increase the level of abstraction even further and use a domain-specific framework to generate code for FPGA HLS. HIPA$^{cc}$ is a publicly available framework for the automatic code generation of image processing algorithms on GPU accelerators. Starting from a C++ embedded DSL, HIPA$^{cc}$ delivers tailored code variants for different target architectures, significantly improving the programmer's productivity. Recently, HIPA$^{cc}$ was extended to also be able to generate C++ code for the C-based HLS framework Vivado HLS from Xilinx [13]. The design flow of the approach is depicted in Figure 1. A recent addition to HIPA$^{cc}$ is the support for multiresolution applications from image processing and scientific computing. The key contributions in this work are therefore, (a) we show how code for multiresolution structures can be automatically generated for C-based HLS, and (b) we demonstrate the versatility of the approach by presenting two case studies, involving applications from medical image processing and scientific computing. The generated target code is derived from a high-level description for image processing algorithms. Therefore, this work uses the high-level description presented in [9].

## II. Background

In multiresolution processing, a certain data set will be represented on different resolution levels. Starting at full resolution (base), for each consecutive level a more coarse-grained representation of the data set is created, as shown in Figure 2. On each level, the same computational operations can be applied, affecting a different relative region size, without modifying the filter kernel. The recursiveness in multiresolution methods and the high degree of parallelism makes these an ideal target for data streaming-oriented FPGA-acceleration. For HLS it is especially beneficial, that the basic construction steps to traverse multiresolution data and the processing function can be reused at every level and must therefore only be designed once.

As the granularity decreases by a factor of four at every stage from the bottom to the top, the accelerator can be designed for a single clock domain by appropriately reducing the throughput by a factor of four compared to the predecessor stage. An ideal method to achieve this in high-level synthesis is to adapt the pipeline initiation interval (II). Designing for a single clock domain also has the advantage that resource requirements can be reduced by relaxing the performance constraints on the coarse-grained higher levels. For example, a divider that has to process a new value in every clock cycle on the lowest level only has to process a new value in every fourth cycle on the next higher level. To save resources, it can either be adapted to operate at an II of four, or if the algorithm requires more than a single division, it can be shared between computations. A major concern for multiresolution systems on FPGAs are the limited memory resources. As data must be merged at the end of the processing cycle, large amounts of data must be buffered on the lowest level while waiting for data from the higher levels. If the buffers are not sufficiently large the design might not be able to complete the processing cycle. In contrast, the size of the buffer affects block RAM and logic resource usage and should therefore not be set too large. Current mid-range FPGAs, provide sufficient memory resources for data sets of up to one million samples in floating point representation. If larger data sets should be processed the buffers on the lower levels might need to be offloaded to external memory.

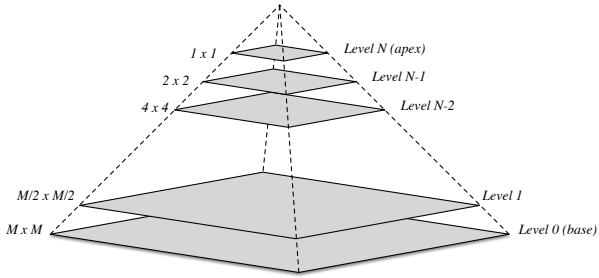

Figure 2. Conceptual representation of multiresolution data.

## III. PROGRAMMING MODEL

The Heterogeneous Image Processing Acceleration (HIPA$^{cc}$) framework consists of a DSL for image processing that is embedded into C++ and a source-to-source compiler. Exploiting the compiler, image filter descriptions written in DSL code can be translated into multiple target languages such as Compute Unified Device Architecture (CUDA), Open Computing Language (OpenCL), Renderscript as used on Android, and C++ code that can be further processed by Vivado HLS [13]. In the following, we will use the Gaussian filter as an example for describing image filters and briefly describe properties of the DSL and show how code generation is accomplished.

*1) Domain-Specific Language:* Embedded DSL code is written by using C++ template classes provided by the HIPA$^{cc}$ framework. The most essential C++ template classes for writing 2D image processing DSL codes are: (a) an Image, which represents the data storage for pixel values; (b) an IterationSpace defining the Region Of Interest (ROI) for operating on the output image; (c) an Accessor defining the ROI of the input image and enabling filtering modes (e. g., nearest neighbor, bilinear interpolation, etc.) on mismatch of

```
1 // input image
2 const int width = 512, height = 512;
3 uchar *image = (uchar*)read_image(width, height, "input.pgm");
4
5 // Gaussian coefficients
6 const float coef[3][3] = { { 0.0625f, 0.1250f, 0.0625f },
7                            { 0.1250f, 0.2500f, 0.1250f },
8                            { 0.0625f, 0.1250f, 0.0625f } };
9
10 Mask<float> mask(coef);
11 Image<uchar> in(width, height);
12 Image<uchar> out(width, height);
13
14 // load image data
15 in = image;
16
17 // reading from in with clamping as boundary condition
18 BoundaryCondition<uchar> bound(in, mask, BOUNDARY_CLAMP);
19 Accessor<uchar> acc(bound);
20
21 // output image
22 IterationSpace<uchar> iter(out);
23
24 // define kernel
25 Gaussian filter(iter, acc, mask);
26
27 // execute kernel
28 filter.execute();
```
Listing 1. Example code for the Gaussian blur filer with kernel size $3 \times 3$.

input and output region sizes; (d) a Kernel specifying the compute function executed by multiple threads, each spawned for a single iteration space point; (e) a Domain, which defines the iteration space of a sliding window within each kernel; and (f) a Mask, which is a more specific version of the Domain, additionally providing filter coefficients for that window. Image accesses within the kernel description are accomplished by providing relative coordinates. To avoid out-of-bound accesses, kernels can further be instructed to implement a certain boundary handling (e. g., clamp, mirror, repeat) by specifying an instance of class BoundaryCondition.

To describe the execution of a Gaussian filter, we need to define a Mask and load the Gaussian coefficients, defined as constants, see Listing 1 (lines 6–10). It is further necessary to create an input and an output image for storing pixel data and loading initial image data into the input image (lines 11–15). The input image is bound to an Accessor with enabled boundary handling mode *clamping* (lines 18–19). After defining the iteration space, the kernel can be instantiated (line 25) and executed (line 28). The actual Kernel implementation is defined elsewhere and not of further importance for the remainder of this work.

In order to describe multiresolution algorithms more efficiently, HIPA$^{cc}$ recently introduced built-in support for image pyramids [12], a common representation of multiresolution data within the domain of image processing [2], which can also be used to describe the multiple scales of data in the multigrid method. To operate on image pyramids, multiple images for the different resolution levels are created, which are then processed by kernels to provide data exchange between these levels (downscaling and upscaling) and between images within the same level. The execution order of those kernels can typically be described in a recursive manner, meaning first downscaling is applied until a certain level, then some operations are executed on one or more of these levels before upscaling is applied to obtain the final image. HIPA$^{cc}$'s language support for image pyramids includes (a) Pyramid, a data structure for automatically creating and encapsulating multiple Images of different resolution; and (b) traverse(), a recursive function embodying the necessary kernel calls for downsampling, upsampling and computations on the same resolution level.

For the multigrid method, data flow is different, as not the input data itself is sampled down (which is called *restriction*), but it is the residual that is calculated and then restricted. However, it is structurally comparable. W-cycles, where—in contrast to the V-cycle—the recursion to the coarser level is carried out twice, can be described by adding an argument to the `traverse()` function call.

*2) Generating Code for Vivado HLS:* The HIPA[cc] compiler is based on the Clang/LLVM 3.4 compiler infrastructure[1]. Utilizing the Clang front end, HIPA[cc] parses C/C++ code and generates an internal Abstract Syntax Tree (AST) representation. Considering Vivado HLS as a target for code generation involves numerous challenges to overcome. Mismatching image sizes in between pyramid levels (e.g., down- and upsampling) need to be handled appropriately, in particular, when transforming the buffer-wise execution model, where kernels are issued one by one, into streaming buffers for pipelining. A pipelined structural description has to be inferred from the linear execution order of kernels. Hereafter, kernel implementations need appropriate placement of Vivado HLS *pragmas* depending on the desired target optimization.

## IV. STREAMING PIPELINE

High-level programs given in HIPA[cc] DSL code process image filters buffer-wise. Each kernel reads from and writes to buffers sequentially, running one after another with buffers serving as synchronization points (so-called host barriers). Buffers can be read and written, copied, reused, or allocated only for the purpose of storing intermediate data.

The buffered concept is fundamentally different from streaming data through kernels and processing a computational step as soon as all input dependencies are available. Kernels are therefore interconnected with each other using stream objects implementing First In First Out (FIFO) semantics. This streaming concept requires a structural description, resolving direct data dependencies, unconstrained from the exact sequential ordering of kernel executions.

We can transform the buffer-wise execution model into a structural description suitable for streamed pipelining by analyzing the DSL host code, replacing memory transfers by stream objects, and generating appropriate kernel code. Vivado HLS can then be instructed to run all kernels in parallel, which can deliver a significantly shorter processing time.

### A. Generating the Pipeline

DSL code is translated into an AST representation that is traversed by HIPA[cc]. During this traversal process, we track the use of buffer allocations, memory transfers and kernel executions by detecting compiler-known classes. For each kernel, the direct buffer dependencies are analyzed and fed into a dependency graph.

Given this graph, we can build up our internal representation, a simplified AST-like structure based on a bipartite graph consisting of two vertex types: *Space* representing buffers and *process* marking kernel executions. By traversing the kernel executions in the sequential order, in which they are specified, writes to buffers are transferred to the internal representation in Static Single Assignment (SSA) manner. Hereby, reused buffers will form new *space* vertices in the graph. Furthermore, when the inputs of multiple kernels depend on the same buffer and the same temporal instance of intermediate data, it is required to replace these dependencies by a *process* for splitting the data, followed by multiple *spaces*, one for each kernel. This way, it is guaranteed that streaming data later on will be copied before handing it over to the next computation steps.

[1] http://clang.llvm.org

Similar considerations need to be taken into account for filtering, which is applied on mismatch of `IterationSpace` size and `Accessor` size in order to match buffer accesses. For the Vivado HLS target, unfiltered access, nearest neighbor, and bilinear filtering are supported by HIPA[cc]. The size discrepancy must be a factor based on a power of two, so that every value of the more coarse-grained levels matches exactly an integral number of input values. Considering multiple levels of multiresolution data, processing takes place within the same level and in between levels. Here as well, *processes* for splitting data into multiple *spaces* must be inserted in order to distribute data among kernels of the same and more coarse-grained levels. Furthermore where filtering is applied, an additional filtering *process* needs to be inserted into the internal representation.

From the internal representation, we can infer the structural description for the streaming pipeline. Every *process* vertex is translated to a kernel execution and every *space* vertex marks the insertion of a unique Vivado HLS stream object for code generation. The resulting code embodies the structural description of the filter, which is written to a file serving as *entry function*.

Kernels described within HIPA[cc]'s language structures for multiresolution methods need to be generated appropriately for each level. This mainly implies reducing resolution and consequently reducing sizes of necessary line and window buffers. Furthermore, to match the latency of the lower levels (coarse-grained), which are processing much less data, with the latency of the higher levels (fine-grained), *pragmas* must be inserted for instructing Vivado HLS to increase the target II depending on chosen resolution reduction. This leads to an II advance by factor $4$ for a resolution reduction of $2$ in each dimension. Therefore, even though a kernel in a multiresolution algorithm is only specified once in DSL code, there is the necessity to generate a separated version for each resolution level.

The resulting AST is transformed back to source code by utilizing Clang's *pretty printer* and written to separate files for each kernel. These files will be included by the *entry function*, which already embeds all executions in a structural description.

### B. Parallelization and Design Optimization

A central element of Vivado HLS for achieving different design goals are synthesis directives, which allow to specify how the input design is to be parallelized and optimized. Synthesis directives in Vivado HLS can either be inserted in the code directly as *pragmas*, or collected in a script file which is applied during synthesis. Apart from several optimization techniques included by HIPA[cc] to improve synthesis results, such as optimizing loop counter variables using assertions and keeping a unified iterations space throughout the designs (refer to [14]), multiresolution applications require to set synthesis directives to control the pipeline rates and the buffer sizes. To allow for changes during implementation, such as to decrease the throughput of the system to reduce resource requirements, the pipeline rate directives are assembled in a script file, as to not require the designer to search through the code. In contrast, appropriate buffer sizes must be defined manually.

## V. CASE STUDIES

We evaluate our methodology on two different hardware target platforms, an embedded General Purpose GPU (GPGPU) (ARM Mali T604) and a mid-range FPGA (Xilinx Zynq 7045). The evaluated designs are compared in terms of performance. The implementations are generated by HIPA$^{cc}$ for each target, stemming from the same code base. The generated code for HLS is synthesized using Vivado HLS 2014.1. The resulting RTL description in VHDL is synthesized, placed and routed using Vivado 2014.1. Power values for the FPGA designs were obtained using Vivado power analysis with toggle information from PPnR netlist simulations using Mentor Graphics QuestaSim.

For the evaluation, we consider two typical multiresolution applications: First an image pyramid based on the Gaussian pyramid, performing a bilateral filtering on different resolution levels and second, a multigrid algorithm for High Dynamic Range (HDR) compression that has been used as a HIPA$^{cc}$ showcase in previous work [10]. These applications greatly demonstrate both the flexibility of HIPA$^{cc}$'s expressiveness and the possibility of target-independent code generation for non-trivial algorithms. Although these algorithms are well known, their implementation details may differ significantly, thus we briefly clarify the algorithm specifics used for our evaluation.

### A. Multiresolution Image Processing

*1) Gaussian Pyramid:* Image pyramids, as depicted in Figure 2, are a fundamental concept in multi-rate image processing. A well-known example is the Gaussian pyramid, which is made up of low-pass filtered, downsampled images of the preceding stage of the pyramid, where the base stage $\mathbf{g}_0$ is defined as the original image $\mathbf{g}_0(\mathbf{x}) = \mathbf{f}(\mathbf{x})$. Higher stages are defined by $\mathbf{g}_s(\mathbf{x}) = \sum_\xi \mathbf{g}_{s-1}(\mathbf{x}) w(\mathbf{x}, \xi)$, where $w(\mathbf{x}, \xi)$ is a weighting function that is identical for all stages, termed the *generating kernel*. The weighting function closely resembles the *Gaussian* function, hence the name of the pyramid. Most practical approaches, however, stop before reaching the top of the pyramid. After processing the images at each stage, the output is reassembled by fusing together images of successive stages in a reconstruction step. For this, the smaller image is first increased in size to match the larger image then the two images are added together. The two basic building blocks for the Gaussian pyramid are downsampling, which we refer to a *decompose*, as well as upsampling and image fusion, which will be referred to as *reconstruct*.

*2) Bilateral Filter:* On the same level within the Gaussian pyramid, we apply the bilateral filter, a non-linear image filter for reducing noise and preserving edges at same time [17]. It is based on a local operator containing two Gaussian filter kernels, $G_{\sigma_s}$ for taking spatial similarity into account, and $G_{\sigma_r}$ for considering range similarity (intensity).

$$I'(x) = \frac{1}{W_p} \sum_{x_i \in \Omega} G_{\sigma_s}(\|x_i - x\|) G_{\sigma_r}(|I(x_i) - I(x)|) I(x_i)$$

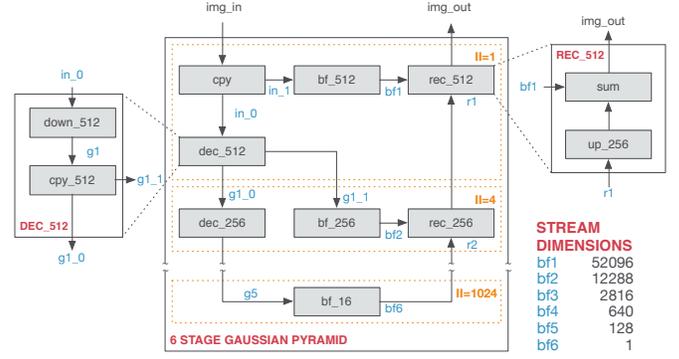

Figure 3. Structural representation of the Gaussian pyramid design.

$$W_p = \sum_{x_i \in \Omega} G_{\sigma_s}(\|x_i - x\|) G_{\sigma_r}(|I(x_i) - I(x)|)$$

The result of both kernels represent the weights for each neighboring pixel $x_i$ contributing to a weighted sum. The second kernel greatly degrades the weight for neighboring pixels with profound difference in intensity. These pixels only contribute very little to the output and thereby it is ensured that edges are preserved.

For evaluation of the multi-rate bilateral filter, we use 8-bit greyscale images of size $512 \times 512$. Basic building blocks for pyramid traversal (*decompose* and *reconstruct*) are implemented using 8-bit unsigned integer arithmetic. The kernel computations are carried out using single precision floating point operators. A structural representation of the multi-rate system is illustrated in Figure 3. The most challenging buffers interconnect the bilateral filter with the reconstruct block. Before reconstruct can start consuming pixels from the buffer, it must wait until the preceding reconstruct module on the next lower level has started to produce pixels. Depending on the pyramid level, the buffer may become very large, as for example on the top level, the buffer must hold over 50,000 entries before data can be consumed. We have evaluated the resource requirements and achievable performance of the design for bilateral filter kernels of size $3 \times 3$ and $5 \times 5$, shown in Table I. The presented results indicate the tremendous amount of logic resources necessary for the computation of the kernels.

### B. Multiresolution in Scientific Computing

*1) Smoother:* In numerical analysis, a smoother is a method to damp high frequencies. In the sense of multigrid methods, smoothers are used to reduce high-frequency components of the error that arises when approximating the solution of PDEs. Commonly used smoothers include the Jacobi method as well as the Gauss-Seidel method. Both methods per se are iterative solvers of linear systems of equations $Ax = b$ and work by calculating a new approximation $x^{(m+1)}$ from the previous approximation $x^{(m)}$, the matrix $A$ and the right-hand side $b$. In case of the Jacobi smoother, the calculation of the new approximation of each component of $x^{(m+1)}$ is independent from other components, whereas for the Gauss-Seidel method, calculation of components depends (in part) on components from the current iteration $(m+1)$. By introducing a relaxation parameter $\omega$ to improve convergence rates, the JOR (Jacobi over-relaxation), respectively SOR (Successive over-relaxation), methods are created.

Table I. PPNR RESULTS OF 6 STAGE MULTIRESOLUTION APPLICATIONS FOR A XILINX ZYNQ 7045.

|        | II | LAT     | SLICE  | LUT     | FF      | BRAM | DSP | F[MHz] | P[W] |
|--------|----|---------|--------|---------|---------|------|-----|--------|------|
| BF 3x3 | 1  | 270 533 | 20 419 | 56 978  | 63 768  | 70   | 368 | 188.1  | 1.9  |
| BF 5x5 | 1  | 296 924 | 41 737 | 113 656 | 126 214 | 76   | 825 | 141.5  | 3.6  |
| Jacobi | 1  | 270 455 | 18 866 | 54 413  | 68 943  | 372  | 259 | 154.3  | 1.4  |

*2) Multigrid Methods:* In scientific computing, multigrid methods are a popular choice for the solution of large systems of linear equations that may stem from the discretization of Partial Differential Equations (PDEs). One of the most popular PDEs is the Poisson equation in order to model diffusion processes. It is similar to the equation to be solved in HDR compression, which is explained in detail in [10].

The V-cycle, a simple scheme of a multigrid method, is shown in Algorithm 1. In the pre- and post-smoothing steps, high-frequency components of the error are damped by smoothers such as the *Jacobi* or the *Gauss-Seidel* methods. In the algorithm, $\nu_1$ and $\nu_2$ denote the number of smoothing steps that are applied. Low-frequency components are transformed into high-frequency components by restricting them to a coarser level, thus making them good targets for the smoother.

On the coarsest level, a direct solution of the remaining linear system of equations is possible due to its low number of unknowns. However, it is also possible to apply a number of smoother iterations. In the case of a single unknown, one smoother iteration corresponds to the direct solution.

**if** *coarsest level* **then**
  solve $A^h u^h = f^h$ exactly or by many smoothing iterations
**else**
  $\bar{u}_h^{(k)} = \mathcal{S}_h^{\nu_1}\left(u_h^{(k)}, A^h, f^h\right)$ {pre-smoothing}
  $r^h = f^h - A^h \bar{u}_h^{(k)}$ {compute residual}
  $r^H = R r^h$ {restrict residual}
  $e^H = V_H\left(0, A^H, r^H, \nu_1, \nu_2\right)$ {recursion}
  $e^h = P e^H$ {interpolate error}
  $\tilde{u}_h^{(k)} = \bar{u}_h^{(k)} + e^h$ {coarse grid correction}
  $u_h^{(k+1)} = \mathcal{S}_h^{\nu_2}\left(\tilde{u}_h^{(k)}, A^h, f^h\right)$ {post-smoothing}
**end**

Algorithm 1. Recursive V-cycle: $u_h^{(k+1)} = V_h\left(u_h^{(k)}, A^h, f^h, \nu_1, \nu_2\right)$.

Figure 4 shows a structural representation of the FPGA implementation of the multigrid HDR compression generated by HIPA$^{cc}$. Here, *restriction* and *prolongation* are the building blocks for grid traversal. The *smoother* is implemented using the JOR method. Throughout the design, we use single precision floating point arithmetic. Hardware resource requirements as well as performance and power results for a design starting from a grid of size $5 \times 5$ are given in Table I. Since the Jacobi smoothers require much less computational complexity than the bilateral filter, it is possible to instantiate many of these kernels for smoothing without stressing logic resource requirements. However, as these are local operators, they require multiple image lines as input before being able to produce output data. Thus, the required buffer sizes to interconnect modules on different sides of the V-cycle are much larger than for the Gaussian pyramid and having to store floating point values severely impacts the amount of required BRAM resources.

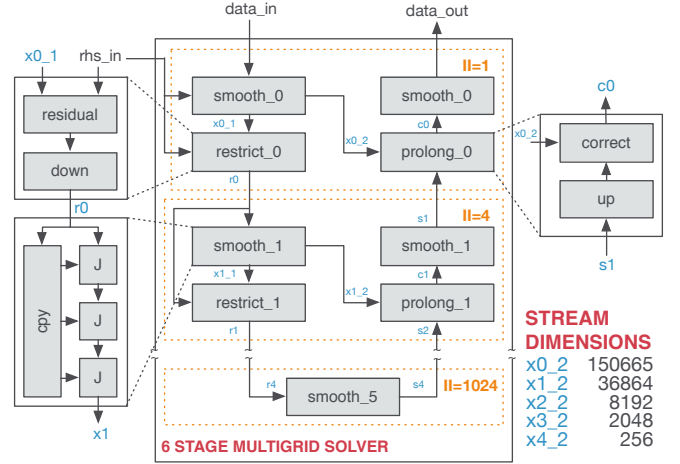

Figure 4. Structural representation of the HDR compression design.

### C. Comparison

As HIPA$^{cc}$ can generate code for several different hardware accelerators, we can use the same code base to compare the performance results of the FPGA implementations to an embedded GPGPU in terms of maximum achievable framerate in frames per second (FPS). As the results presented in Table II show, the high degree of parallelism of the FPGA can be fully exploited to achieve a very high framerate in comparison to the ARM Mali. Performance results for the ARM Mali are mainly restrained by memory bandwidth, and therefore strictly depend on the number of memory accesses, defined by the chosen window size. This is clearly demonstrated as the achievable framerates for the bilateral filter deviate approximately by a factor of $9/25$.

Table II. ACHIEVABLE FRAME RATES IN FRAMES PER SECOND (FPS) FOR MULTIRESOLUTION APPLICATIONS PROCESSING 6 RESOLUTION LEVELS STARTING AT $512 \times 512$.

|              | Mali T604 | Zynq 7045 |
|--------------|-----------|-----------|
| BF $3 \times 3$ | 54.35     | 695       |
| BF $5 \times 5$ | 19.73     | 476       |
| Jacobi       | 37.11     | 570       |

## VI. RELATED WORK

Numerous HLS approaches, both in academia and industry, have been developed over the past decades. Most of them start from a simplified imperative programming language, e. g., a subset of C, which is translated by stepwise refinement into a synthesizable Hardware Description Language (HDL) description—Calypto's Catapult, Forte's Cynthesizer, the Synphony C Compiler from Synopsys, or Vivado HLS from Xilinx are the most well-known commercial approaches. For a specific field of application, it is often a challenge to bring together different areas of expertise, for instance, mathematics, algorithm engineering, and parallel code or hardware generation. One way to productivity are programming abstractions, such as libraries or DSLs. In the domain of image processing, HLS frameworks sometimes include specific libraries to provide elemental architecture constructs and filtering implementations, as for example, the partial port of the OpenCV library [1] for Vivado HLS from Xilinx [18] or the smart buffer concept in ROCCC [4]. Extending such libraries might become quite a burden and lowers portability to new target architectures. In contrast, DSL-based approaches decouple the algorithm specification from the implementation and hardware details. They are much for flexible and can be easily extended to generate code for different platforms. PARO [5], for instance, is a HLS environment that provides domain-specific augmentations [15] for image processing (e. g., border treatment and reductions such as median filtering), it has been successfully used for adaptive multiresolution filtering in medical imaging [6]. Another recent approach that can emit parallel code for multi-core systems as well as generate hardware pipelines was proposed by Hegarty et al. [7]. But, none of the two aforementioned approaches offers support at language level for image pyramids. For stencil computations there exist several DSL-based approaches [3, 16, 8], however, they consider hardware specifics only to a limited extend, and target only multi-core systems and GPU but not FPGA accelerators. To the best of our knowledge, our approach is the first one that can generate performance-portable code for GPUs as well as HDL code for multiresolution applications.

## VII. Conclusions

In this work, we have demonstrated how the DSL-based framework HIPA$^{cc}$ can be used to automatically generate highly optimized code for the HLS of multiresolution applications for implementation on FPGAs. In this way, the specification of the design requires significantly less programming effort from the developer and thus also poses less chances for coding errors. The presented case studies from medical image processing and scientific computing demonstrate that the approach is applicable to broad range of multiresolution problem scenarios. As HIPA$^{cc}$ also includes embedded GPGPUs as a hardware target [11], we have compared the proposed FPGA approach to a highly optimized GPU implementations, generated from the same code base. The assessment exposes the benefits of using a heterogeneous framework for algorithm development and can easily identify a suitable hardware target for efficient implementation.

## Acknowledgment

We thank Richard Membarth for providing the tested multi-grid implementation in HIPA$^{cc}$. This work was partly supported by the German Research Foundation (DFG) as part of the Research Training Group 1773 "Heterogeneous Image Systems" and as part of the Priority Programme 1648 "Software for Exascale Computing" in project under contract TE 163/17-1.